\documentclass{article}
\usepackage{spconf,amsmath,graphicx}

\usepackage{kotex}
\usepackage{booktabs}
\usepackage{kotex}
\usepackage{xcolor}
\usepackage{multirow}
\usepackage{xspace}
\usepackage{enumitem}

\let\OLDthebibliography\thebibliography
\renewcommand\thebibliography[1]{
  \OLDthebibliography{#1}
  \setlength{\parskip}{0pt}
  \setlength{\itemsep}{0pt plus 0.5ex}
}

\graphicspath{{Figures/}}
\DeclareGraphicsExtensions{.pdf,.png,.jpg}

\definecolor{darkgreen}{rgb}{0.1, 0.8, 0.1}
\definecolor{orange}{rgb}{0.8, 0.5, 0.1}
\definecolor{blue}{HTML}{3282b8}

\newcommand\axlm{ST-BERT\xspace}
\title{\axlm: Cross-modal Language Model Pre-training \\ for End-to-end Spoken Language Understanding}
\name{Minjeong Kim$^1$, Gyuwan Kim$^1$, Sang-Woo Lee$^{1,2}$, Jung-Woo Ha$^{1,2}$}
\address{$^1$NAVER CLOVA, $^2$NAVER AI LAB}
\begin{document}
%
\maketitle

\begin{abstract}
Language model pre-training has shown promising results in various downstream tasks. In this context, we introduce a cross-modal pre-trained language model, called Speech-Text BERT (\axlm{}), to tackle end-to-end spoken language understanding (E2E SLU) tasks. Taking phoneme posterior and subword-level text as an input, ST-BERT learns a contextualized cross-modal alignment via our two proposed pre-training tasks: Cross-modal Masked Language Modeling (CM-MLM) and Cross-modal Conditioned Language Modeling (CM-CLM). Experimental results on three benchmarks present that our approach is effective for various SLU datasets and shows a surprisingly marginal performance degradation even when 1\% of the training data are available. Also, our method shows further SLU performance gain via domain-adaptive pre-training with domain-specific speech-text pair data.

\end{abstract}
\begin{keywords}
Cross-modal language model, Cross-modal pre-training, Spoken language understanding
\end{keywords}

\section{Introduction}

End-to-end spoken language understanding (E2E SLU) has recently shown promising results~\cite{serdyuk2018towards,lugosch2019speech,wang2020large}. They outperform the previous SLU approaches that cascade automatic speech recognition and natural language understanding (ASR-NLU) for intent classification for the spoken utterances.
The main advantage of E2E SLU models is their ability to understand the speaker's intent without translating the speech to a text transcript.
This allows the models to fully exploit additional information such as emotion and nuance characterized with acoustic signals. 
Recently, leveraging large-scale pre-trained language models (PLMs) such as BERT~\cite{devlin2019bert} has enhanced SLU performances~\cite{cho2020speech, denisov2020pretrained} by benefiting from richly learned textual representation. However, these methods exploit only limited textual information by explicitly aligning the spoken utterance and its transcript representations. Hence, existing E2E SLU methods can be further improved in terms of effective learning of PLM-based speech and text representations.
 
\begin{figure}
\centering 
\includegraphics[width=0.48\textwidth]{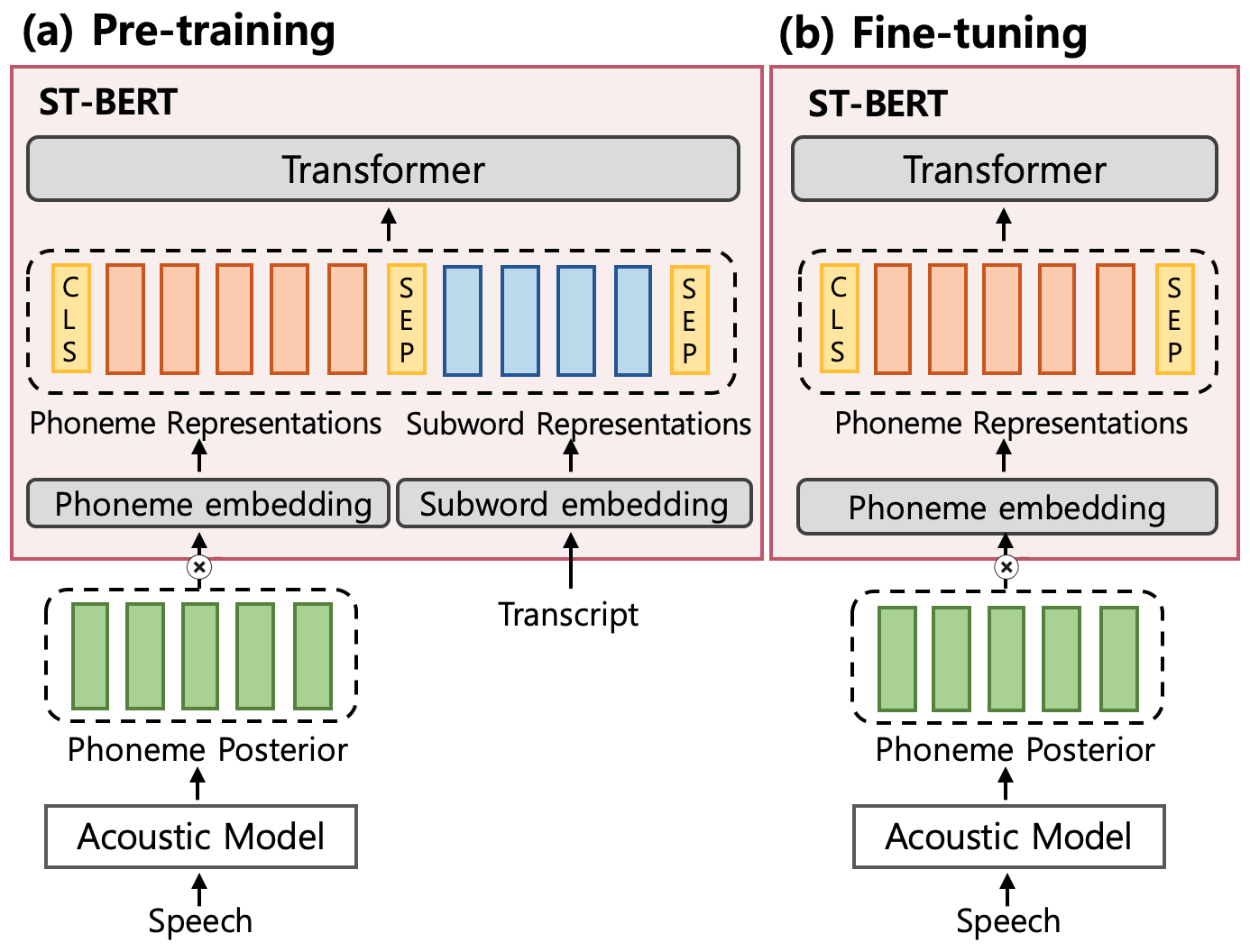}
\vspace*{-7mm}
\caption{Overall pre-training and fine-tuning procedures for \axlm{}. \label{fig:model}
}
\vskip -0.1in
\end{figure} 
Here we propose a new E2E SLU method that can effectively leverage both speech and text modalities based on a BERT-based PLM, called Speech-Text BERT (\axlm). We extended our model with an acoustic model to accommodate speech data. Training \axlm consists of two steps: pre-training from large-scale corpora such as Librispeech~\cite{panayotov2015librispeech} and fine-tuning for the downstream tasks. We define two new pre-training methods, Cross-modal Masked LM (CM-MLM) and Cross-modal Conditional LM (CM-CLM), 
inspired by the recent success of cross-lingual translation LM research~\cite{lample2019cross}. By masking a whole or a part of a sequence of each modality and predicting the masked sequence from the other modality, \axlm can learn the implicit speech-text alignment and utilize the rich representation of a PLM, thus improving SLU performances. 


Our contributions are summarized in three folds: 
\begin{itemize}[noitemsep,topsep=0pt,parsep=0pt,partopsep=0pt]
    \item We propose a new cross-modal pre-trained LM-based SLU method, \axlm, which learns contextualized speech-text alignment by defining two new cross-modal LM pre-training tasks.
    \item We demonstrate domain-adaptive pre-training (DAPT) for SLU tasks when 
    transcripts of downstream task data are available and show DAPT can further improve SLU performance when being applied to \axlm.
    \item The proposed \axlm shows state-of-the-art performance on three SLU datasets, especially achieving large performance gains under data shortage scenarios~with extensive experiments and ablation studies.
\end{itemize}

\label{sec:intro}

\section{Related Works}

\subsection{E2E SLU}
Many E2E SLU methods have been proposed to address the limitations of ASR-NLU approaches~\cite{lugosch2019speech,wang2020large,tomashenko2019recent}.
Serdyuk et al. \cite{serdyuk2018towards} follow an E2E ASR architecture for E2E SLU
but the model is not competitive compared to the ASR-NLU pipeline.
Instead of naive supervised learning from the labeled SLU data, recent approaches incorporate speech model pre-training which is crucial to achieving successful accuracy, especially when the number of labeled data is scarce.
They first pre-train an audio encoder with auxiliary tasks (e.g., ASR) and then fine-tune it after adding additional layers for target SLU tasks \cite{lugosch2019speech,chen2018slu,bhosale2019end}.
For instance, Lugosch et al. \cite{lugosch2019speech} pre-train a SincNet-based neural network \cite{ravanelli2018speaker} with phonemes and words base ASR task and fine-tune it for intent classification after adding additional recurrent layers.

\subsection{Cross-modal Representation Learning}

Inspired by cross-lingual language models (XLMs) \cite{lample2019cross}, we interpret two modalities, speech, and text in our case, as a different language to represent shared meaning. 
We borrow pre-training techniques from \cite{lample2019cross} to jointly train a cross-modal language model.
Cross-modal pre-training also draws huge attention in the computer vision community \cite{li2020unicoder,lu2019vilbert} to transfer learned knowledge to vision-and-language tasks \cite{antol2015vqa,zellers2019recognition}.

Chung et al. \cite{chung2018unsupervised} learn audio segment representations and word representations individually and aligns their spaces via adversarial training.
Other works \cite{cho2020speech,denisov2020pretrained} match sequence-level representations of the two modalities using knowledge distillation \cite{hinton2015distilling} from a text encoder to a speech encoder.
SpeechBERT \cite{chuang2019speechbert} jointly trains multi-modal representations.
However, the inputs from two different modalities are semantically different, thus loosely connected. 
This circumstance makes pre-training less effective though it is trained with masked language modeling like BERT \cite{devlin2019bert}.

\begin{figure*} 
\centering 
\includegraphics[width=0.82\textwidth]{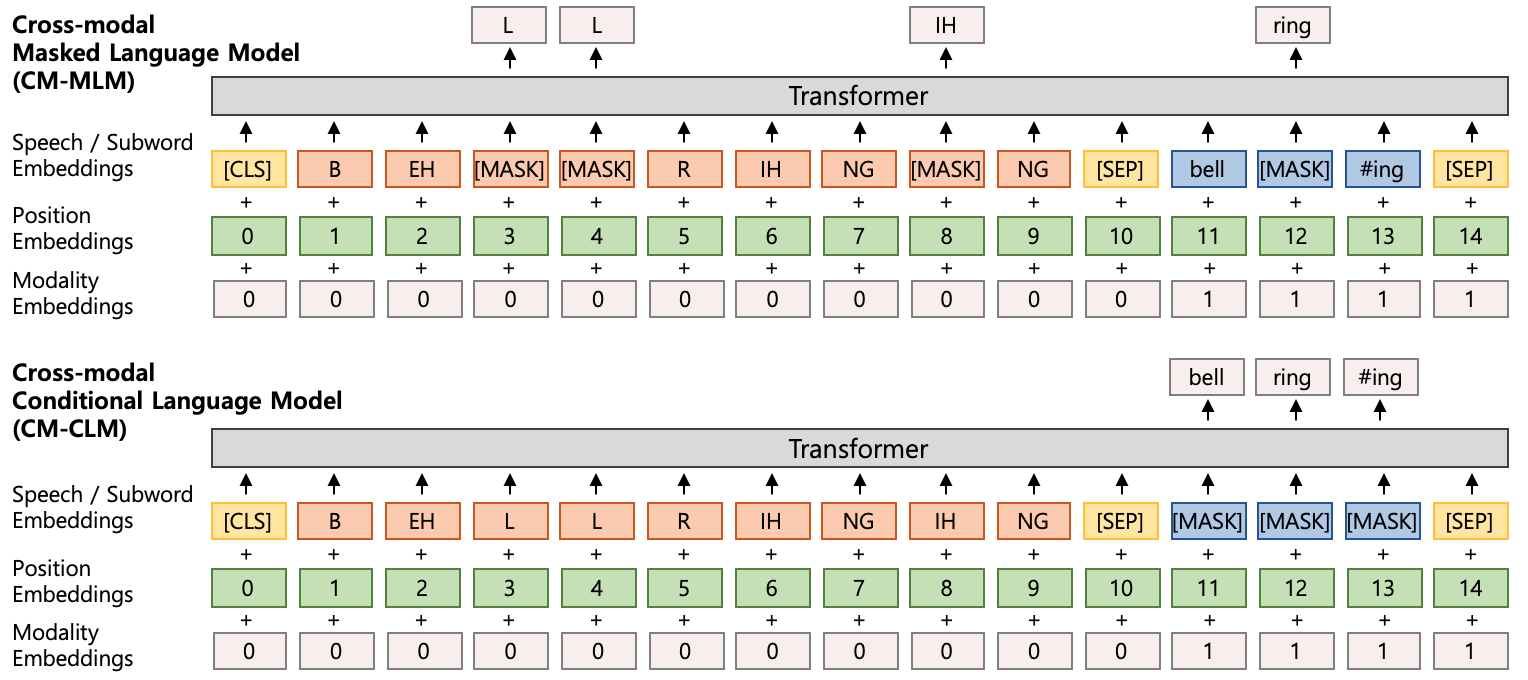}
\vspace*{-3mm}
\caption{Cross-modal language modeling objectives for \axlm{} pre-training. We illustrate speech-to-text CM-CLM to represent CM-CLM.\label{fig:pretrain}
}
\vskip -0.1in
\end{figure*} 

\section{Proposed Method}

\subsection{Model Architecture} 
\axlm{} is a cross-modal extension of BERT \cite{devlin2019bert} that can handle both speech data $x^s$ and text data $x^t$ for SLU.
We incorporate a pre-trained acoustic model (AM) to accommodate speech data as an input to the BERT as shown in Fig. \ref{fig:model}.

The AM estimates 
the posterior distribution over phonemes from a raw waveform. We follow the model structure and its ASR pre-training procedure as presented in~\cite{lugosch2019speech}. In specific, the AM consists of a phoneme module and a word module. The former is based on a SincNet layer followed by convolutional and recurrent layers and the latter employs recurrent layers. Each module learns to predict phoneme and word targets.

After ASR training, we discard the word module and combine the phoneme module only with ST-BERT.
Pre-trained AM remains frozen during the ST-BERT pre-training and fine-tuning to extract generic acoustic features.   

\axlm takes phoneme posterior and subword-level tokenized text as an input and encodes them into phoneme and subword embeddings, respectively.
A subword embedding is obtained via looking up a subword embedding matrix, whereas a phoneme embedding is computed as a weighted sum of phoneme embedding vectors based on the phoneme posterior. The input sequence starts with the embedding of a special classification token ($[CLS]$) and the embeddings of each modality follow with that of a special separator token ($[SEP]$) at the end of each sequence. 
The summation of speech (or subword), modality, and position embeddings is fed into the transformer encoder as an input for each time step as in Fig. \ref{fig:pretrain}.

It is worth noting that our transformer encoder takes the concatenation of phoneme and subword representations during pre-training (Fig. 1-(a)), while it utilizes phoneme representations only for fine-tuning (Fig. 1-(b)).
Also, we initialize \axlm, except for the phoneme embedding layer, with pre-trained BERT to leverage rich textual information.

\subsection{Pre-training}
We propose two cross-modal language modeling objectives to learn contextualized cross-modal alignments between speech and text.
Both objectives require speech-text pairs and take their concatenation as an input, as illustrated in Fig. \ref{fig:pretrain}. 

\subsubsection{Cross-modal Masked Language Modeling}
Cross-modal Masked Language Modeling (CM-MLM) is similar to the translation language modeling task which is suggested for cross-lingual pre-training \cite{lample2019cross, huang2019unicoder}. We randomly sample 15\% of input elements and replace them with [MASK] token embedding. The model is trained to predict a masked phoneme or subword by aggregating the information from the representations of the same modality and its pair components in the other modality. This encourages the model to implicitly learn the contextualized alignments between both modalities. To prevent the model from easily predicting a masked representation from its neighbor, when masking phonemes, we mask out the phoneme embedding span corresponding to the same target phoneme.
The training objective of CM-MLM is 
$L_{M}=-\log p(x^s_m, x^t_m|\bar{x^s}, \bar{x^t})$,
where $\bar{x}$ is a masked sequence and $x_m$ is a sequence of original tokens for masked positions, for each modality.

\subsubsection{Cross-modal Conditioned Language Modeling}
Cross-modal conditioned language modeling (CM-CLM) is a new task to more effectively learn speech-text cross-modal relationships.
Instead of masking the input sequence with a possibility of 15\%, we mask the entire sequence of target modality and train the model to predict the masked representations solely conditioned on the source modality. 
Thus, there exist two types of CM-CLM: speech-to-text CM-CLM and text-to-speech CM-CLM. The former aims to predict masked text tokens given speech and the latter predicts vice versa. 

When inferring the masked phoneme representation, in CM-MLM task, it is likely that a model simply relies on the context of phoneme sequence neglecting the textual information. However, CM-CLM task forces the model to exploit nothing but a given text by its design. Hence, CM-CLM is a more challenging pre-training task in learning implicit cross-modal alignment and has proven to be effective in our experiments.
The training objectives of speech-to-text CM-CLM and text-to-speech CM-CLM are
$L^T_{C}=-\log p(x^t|x^s, m^t)$ and $L^S_{C}=-\log p(x^s|m^s, x^t)$ respectively,
where $m^t$ (or $m^s$) is a sequence of $[MASK]$ with the length of $x^t$ (or $x^s$).

\subsection{Fine-tuning}
Fine-tuning ST-BERT for SLU is straightforward. A phoneme posterior extracted from task-specific raw waveform via AM is fed into ST-BERT along with $[CLS]$ and $[SEP]$ token embeddings. Following the typical BERT-based fine-tuning strategy, the final hidden states of $[CLS]$ token is fed into a linear classifier on top of the pre-trained \axlm{}. As in pre-training, all the parameters except AM are fine-tuned.  

\subsection{Domain-Adaptive Pre-training}
When domain-specific speech-text pair data are available in addition to the intent annotations, domain-adaptive pre-training (DAPT) \cite{gururangan2020dont} is likely to improve the fine-tuning performance. We simply perform additional \axlm pre-training on downstream task data with transcripts before fine-tuning. 
\section{Experiment}

\subsection{Implementation} \label{training_detail}
\emph{\textbf{Pre-training. }}
We pre-trained both the acoustic model and \axlm{} on 960 hours of Librispeech \cite{panayotov2015librispeech}. We use publicly available speech and text alignment\footnote{https://zenodo.org/record/2619474\#.X413GEIzZds}  \cite{lugosch2019speech} which is obtained using Montreal Forced Aligner \cite{mcauliffe2017montreal}. For the text input, we use WordPiece tokenizer \cite{wu2016google} to segment the text. 

We adopt a curriculum pre-training \cite{wang2020curriculum} for better generalization and fast convergence \cite{bengio2009curriculum}. \axlm{} pre-training starts with CM-MLM task, the easier task, for the one-third of the entire training. After then, one of CM-MLM, speech-to-text and text-to-speech CM-CLM tasks is sampled with an equal probability for every iteration during the rest of pre-training. 

We initialize the model parameters from BERT-base and use Adam \cite{kingma2014adam} optimizer with linear-decay learning rate schedule where a peak learning rate is 1e-4. We train the model for about 160K steps with a 512 size mini-batch where the maximum input sequence length is set to 256. 

\medskip
\noindent\emph{\textbf{Fine-tuning. }} 
We mainly evaluate our \axlm{} on the Fluent Speech Command (FSC) dataset \cite{lugosch2019speech}, one of the most widely used datasets for the E2E SLU tasks. 
This dataset includes 23,132 training utterances of 77 speakers, 3,118 validation utterances of 10 speakers, and 3,793 test utterances of 10 speakers. It contains 31 unique intents.

Additionally, our model is validated on the Snips and SmartLight datasets for intent classification tasks.
Snips dataset \cite{coucke2018snips} is an NLU benchmark, which consists of utterances and their corresponding intent-slot labels.
The number of training, validation, and test data is 13,084, 700, and 700 utterances respectively, with 7 unique intents. 
Previous SLU studies used this dataset by synthesizing audio from text data, and we follow the same experimental protocol. To compare our result with the result of Huang and Chen, we used Google text-to-speech (TTS) system the same as in \cite{huang2020learning}, though it does not guarantee that our data are identical to their audio.

SmartLights dataset \cite{saade2018spoken} consists of 1,660 spoken commands for a smart light assistant with 6 unique intents. 
For each utterance, there are two audio types where the microphone setting is different, \textit{close field} and \textit{far field} and the latter setting is more challenging. To evaluate our model on this dataset, we use 10-fold cross-validation as suggested in \cite{huang2020learning}.

In our full data experiments, we measure a model performance twice. For the 10\% and 1\% experiment in Table \ref{table:fsc}, we evaluate our model on 10 and 20 random subsets, respectively. The reported numbers are the average score. We fine-tune our model with a maximum sequence length of either 128 or 256 depending on the dataset and set a batch-size to 32. All the experiments were performed based on NAVER Smart Machine Learning (NSML) platform~\cite{kim2018nsml}.


\subsection{Results}

\begin{table}[t]
\begin{center}
\caption{Test accuracies on the FSC dataset. 10\% and 1\% denotes the data shortage scenarios where 10\% and 1\% of the training data are given for training.}
\label{table:fsc}
\small
\begin{tabular}{p{3.8cm} r r r}
\\ \toprule 
Model & \multicolumn{1}{c}{Full} & \multicolumn{1}{c}{10\%} & \multicolumn{1}{c}{1\%} \\ 
\midrule 
Lugosch et al. \cite{lugosch2019speech} & 98.80\% & 97.96\% & 82.78\%  \\
Wang et al.  \cite{wang2020large} (BERT)  & 98.95\% & - & -  \\
Wang et al. \cite{wang2020large} (ERNIE)  & 99.02\% & - & - \\
Cho et al. \cite{cho2020speech} & 98.98\% & 98.12\% & 83.12\%  \\
Price \cite{price2020end} & 99.3\% & - & -  \\
\midrule
\textbf{\axlm (Ours)} & \textbf{99.50\%} & \textbf{99.13\%}  & 95.64\%  \\
\hline
- CM-CLM & 99.42\% & 99.09\%  & \textbf{96.51\%}  \\
- text data (pre-training) & 99.39\%  & 99.04\%  & 89.81\%  \\
\hline
+ DAPT & \textbf{99.59\%} & \textbf{99.25\%}  & 95.83\% \\
\bottomrule
\end{tabular}
\end{center}
\vskip -0.2in
\end{table}

\emph{\textbf{Main Results.~} }
Tables \ref{table:fsc} and \ref{table:smartlights_snips} show that our model outperforms the other comparative models for all datasets. For a fair comparison, we exclude the model performance of a data augmentation setting.
Note that the experiment settings for Smartlights and Snips datasets are not exactly same as that of Huang and Chen~\cite{huang2020learning} due to the given data condition described in \ref{training_detail}. Nevertheless, we present both results to show that our model is competitive.

\medskip \noindent
\emph{\textbf{Ablation Studies. ~}}
We perform the ablation experiments to further investigate the contribution of CM-CLM and a cross-modal pre-train itself.
We compare ST-BERT with 1) ST-BERT pre-trained without CM-CLM task 
and 2) ST-BERT without CM-CLM and text data, i.e. a model pre-trained with MLM task on speech data only. The overall results in Tables \ref{table:fsc} and \ref{table:smartlights_snips} present that both of our approaches are effective to improve model performances. Unlike other datasets, MLM on speech data only shows better accuracy than ST-BERT without CM-MLM for Snips dataset. We conjecture that the usage of the synthesized voice might lead to this result. Nevertheless, our method presents remarkable improvement compared to the baseline.   

%

\medskip \noindent
\emph{\textbf{Data Shortage Scenario.~}}
To examine the robustness of model performance to varying training data size, we test our model with a small amount of data as presented in \cite{lugosch2019speech,cho2020speech}. 
In Table \ref{table:fsc}, we observe a comparatively marginal performance degradation in \axlm{} for both of data shortage scenarios. Especially when only 1\% of data are given, cross-modal pre-training methods lead to a significantly robust performance while uni-modal pre-training on speech suffers from a relatively large decrease in performance. This result demonstrates that leveraging textual information during pre-training is critical when only limited downstream task data are available which is common in reality. 


\begin{table}[t]
\begin{center}
\caption{Test accuracies on the SmartLights and Snips dataset.}
\label{table:smartlights_snips}
\small
\begin{tabular}{p{3.8cm} r r r}
\\ \toprule 
\multirow{2}{*}{Model} & \multicolumn{2}{c}{SmartLights} & \multicolumn{1}{c}{Snips}\\ 
 & \multicolumn{1}{c}{Far} & \multicolumn{1}{c}{Close} & \multicolumn{1}{c}{-} \\
\midrule 
Huang and Chen \cite{huang2020learning} & 47.38\% & 67.98\% & 89.55\% \\
\midrule 
\textbf{\axlm~(Ours)} &\textbf{60.98\%} & \textbf{84.65\%} & \textbf{96.21\%}  \\
\hline
- CM-CLM & 57.06\% & 81.97\% & 95.07\% \\
- text data (pre-training) & 54.93\% &  81.22\% & 95.64\% \\
\hline
+ DAPT & \textbf{69.40\%} & \textbf{86.91\%} &  96.07\%  \\
\bottomrule
\end{tabular}
\end{center}
\vskip -0.1in
\end{table}


\medskip \noindent
\emph{\textbf{Domain-Adaptive Pre-training (DAPT).~}}
Although ST-BERT shows an outstanding performance already without using domain-specific speech-text pair data, applying DAPT to ST-BERT leads to further improvement as presented at the bottom of Tables \ref{table:fsc} and \ref{table:smartlights_snips}. In particular, we confirm that DAPT is remarkably effective to improve the model performance on a more noisy SLU dataset. For example, in Table~\ref{table:smartlights_snips}, ST-BERT attained 8.4\% of accuracy gain in the \textit{far field} of SmartLights dataset which is recorded from 2 meters away.




\section{Conclusion}
We present \axlm, a novel cross-modal pre-training method for SLU tasks, which effectively learns semantic representations of speech and text modalities by defining two new cross-modal language modeling tasks, CM-MLM and CM-CLM. With extensive experiments, we demonstrate that our \axlm not only outperforms state-of-the-art models on three benchmark SLU datasets but also shows robustly better results than baseline models under data shortage scenarios. Also, we apply DAPT to pre-training \axlm, confirming that DAPT can lead to further performance improvement when the transcripts of downstream task data are available.



\clearpage


\bibliographystyle{IEEEbib-abbrev}
\bibliography{refs_simple}

\end{document}